\documentclass[10pt]{article}

\pdfoutput=1

\usepackage[latin1]{inputenc}
\usepackage{mystyle,amssymb,dsfont,amsmath,amsthm,mathtools}

\title{GAN and VAE from an \\ Optimal Transport Point of View}

\author{
  Aude Genevay\footnote{CEREMADE, Univ. Paris-Dauphine, Mokaplan, INRIA \texttt{genevay@ceremade.dauphine.fr}}, 
  \quad
   Gabriel Peyr\'e\footnote{CNRS and DMA, \'Ecole Normale Sup\'erieure,  \texttt{gabriel.peyre@ens.fr}}, 
   \quad
   Marco Cuturi\footnote{ENSAE CREST, Universit\'e Paris-Saclay,  \texttt{marco.cuturi@ensae.fr}}
 }

\date{\today}

\begin{document}

\maketitle

\begin{abstract}
	This short article revisits some of the ideas introduced in~\cite{WassersteinGAN} and~\cite{Bousquet2017} in a simple setup. 
	This sheds some lights on the connexions between Variational Autoencoders (VAE)~\cite{VAE}, Generative Adversarial Networks (GAN)~\cite{GAN} and Minimum Kantorovitch Estimators (MKE)~\cite{bassetti2006minimum,montavon2016wasserstein,bernton2017inference}.
\end{abstract}

\section{Minimum Kantorovitch Estimators}

\paragraph{MKE.} Given some empirical distribution $\nu \eqdef \frac{1}{n}\sum_{j=1}^n \de_{y_j}$ where $y_j \in \Xx\subset\RR^p$, and a parametric family of probability distributions $(\mu_\th)_{\th\in\Theta} \subset \Pp(\Xx)$, $\Theta\subset \mathbb{R}^q$, a Minimum Kantorovitch Estimator (MKE)~\cite{bassetti2006minimum,montavon2016wasserstein,bernton2017inference} for $\th$ is defined as any solution of the problem
\eql{\label{eq-fitting-energy-orig}\tag{MKE}
	\umin{\th} W_c(\mu_\th,\nu), 
}
where $W_c$ is the Wasserstein cost on $\Pp(\Xx)$ for some ground cost function $c:\Xx\times\Xx \rightarrow \RR$, defined as
\eql{\label{eq-primal-formula}
	W_c(\mu,\nu) = \umin{\ga \in \Pp(\Xx \times \Xx)} \enscond{ \int_{\Xx \times \Xx} c(x,y) \d \ga(x,y) }{ P_{1\sharp}\ga=\mu, P_{2\sharp}\ga=\nu },
}
where $P_1(x,y)=x$ and $P_2(x,y)=y$, and $P_{1\sharp}$ and $P_{2\sharp}$ are marginalization operators that return for a given coupling $\gamma$ its first and second marginal, respectively. 

The notations $P_{1\sharp}$ and $P_{2\sharp}$ above agree with the more general notion of pushforward measures: Given a measurable map $g: \Zz\rightarrow \Xx$, which can be interpreted as a function ``moving'' points from a measurable space to another, one can naturally extend $g$ to become a more general map $g_\sharp$ that can now ``move'' an entire probability measure on $\Zz$ towards a new probability measure on $\Xx$. The operator $g_\sharp$ ``pushes forward'' each elementary mass of a measure $\zeta$ in $\Pp(\Zz)$ by applying the map $g$ to obtain then a mass in $\Xx$, to build on aggregate a new measure in $\Pp(\Xx)$ written $g_{\sharp}\zeta$. More rigorously, the pushforward measure of a measure $\zeta \in \Pp(\Zz)$ by a map $g: \Zz\rightarrow \Xx$ is the measure denoted as $g_{\sharp}\zeta$ in $\Pp(\Xx)$ such that for any set $B\subset\Xx$, $(g_{\sharp}\zeta)(B) \eqdef \zeta(g^{-1}(B))=\zeta(\enscond{z\in\Zz}{g(z)\in B})$.

\begin{figure}[ht]
\centering
	\includegraphics[width=\linewidth]{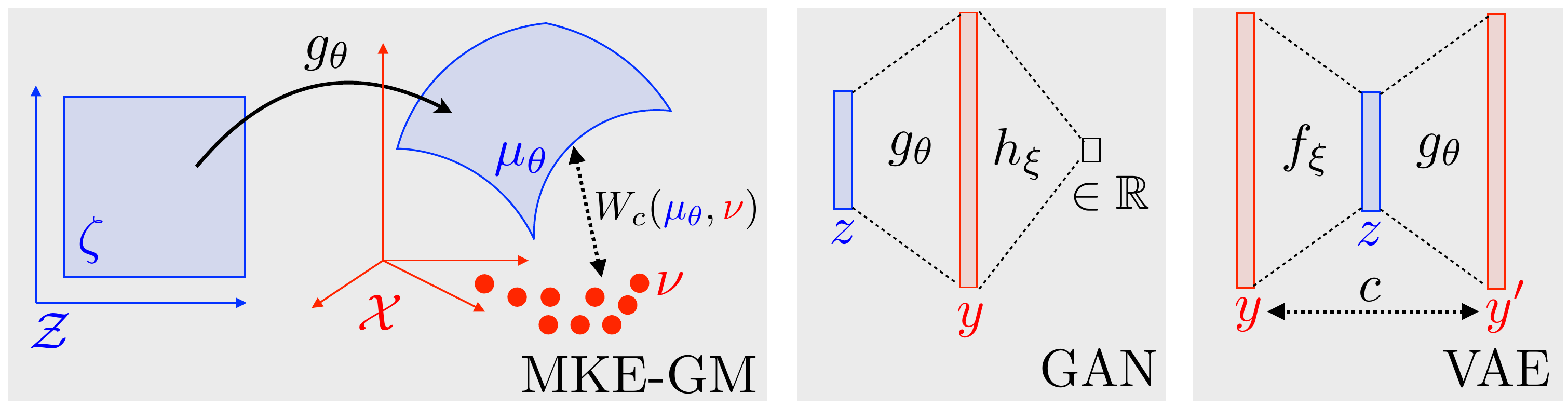}
\caption{\label{fig-workflow} %
	Left: illustration of density fitting using the Minimum Kantorovitch Estimator for a generative model.
	Middle and right: comparison of the GAN vs. VAE setups. 
		}
\end{figure}

\paragraph{MKE-GM.} The MKE approach can be used directly in the case where $(\mu_\theta)_{\theta}$ is a statistical model, namely a parameterized family of probability distributions with a given density with respect to a dominant base measure, as considered for instance with exponential families on discrete spaces in~\cite{montavon2016wasserstein}. However, the MKE approach can also be used in a \emph{generative model} setting, where $\mu_\th$ is defined instead as the push forward of a fixed distribution $\zeta$ supported on a low dimensional space $\Zz\subset\RR^d$, $d\ll p$, where the parameterization lies now in choosing a map $g_\th : \Zz \mapsto \Xx$, i.e. $\mu_\th = g_{\th\sharp}\zeta$, resulting in the following special case of the original~\eqref{eq-fitting-energy-orig} problem:

\eql{\label{eq-fitting-energy}\tag{MKE-GM}
	\umin{\th} E(\th) \eqdef W_c(g_{\th\sharp}\zeta,\nu), 
}
The map $g_\th$ should be therefore thought as a ``decoding'' map from a low dimensional space to a high dimensional space.
In such a setting, the maximum likelihood estimator is in general undefined or difficult to compute (because the support of the measures $\mu_\th$ are singular) while MKEs are attractive because they are always well defined.

\section{Dual Formulation and GAN}

Because~\eqref{eq-primal-formula} is a linear program, it has a dual formulation, known as the Kantorovich problem~\cite[Thm. 5.9]{villani2008optimal}:
\eql{\label{eq-dual-w}
	E(\th) = \umax{h,\tilde h}  
	 \enscond{ \int_{\Zz}  h(g_\th(z)) \d\zeta(z) + \int_{\Xx} \tilde h(y) \d\nu(y)  }{ h(x) + \tilde h(y) \leq c(x,y) }.
}
where $(h,\tilde h)$ are continuous functions on $\Xx$ often called Kantorovich potentials in the literature.

In the dual formulation~\eqref{eq-dual-w}, $\th$ does not appear anymore in the constraints. Therefore, the gradient of $E$ can be computed as
\eql{\label{eq-grad-dual}
	\nabla E(\th) = \int_{\Zz} [\partial_\th g_\th(z)]^\top \nabla h^\star( g_\th(z) ) \d \zeta(z), 
}
where $h^\star$ is an optimal dual function solving~\eqref{eq-dual-w}. Here $[\partial_\th g_\th(z)]^\top \in \RR^{q \times p}$ is the adjoint of the Jacobian of $\th \mapsto g_\th(z)$, where $q$ is the dimension of the parameter space $\Theta$.  

A key remark in Kantorovich's formulation is to notice that the cost of any pair $(h,\tilde h)$ can always be improved by replacing $\tilde h$ in~\eqref{eq-dual-w} by the $c$-transform $h^c$ of $h$ defined as
\eq{
	h^c(y) \eqdef \umax{x} c(x,y) - h(x),
} 
which is, indeed, given a candidate potential $h$ for the first variable, the best possible potential that can be paired with $h$ that satisfies the constraints of~\eqref{eq-dual-w} (see~\cite[Thm. 5.9]{villani2008optimal}). For this reason, one can parameterize problem \eqref{eq-dual-w} as depending on one potential function only. 

A first approach to solve~\eqref{eq-dual-w} is to remark that since $\nu$ is discrete, one can replace the continuous potential $\tilde h$ by the discrete vector $( \tilde h(y_j) )_j \in \RR^n$ and impose $h=(\tilde h)^c$. As shown in~\cite{2016-genevay-nips}, the optimization over $\tilde h$ can then be achieved using stochastic gradient descent.

Similarly to~\cite{WassersteinGAN}, another approach is to approximate~\eqref{eq-dual-w} by restricting the dual potential $h$ to have a parametric form $h=h_{\xi} : \Xx \rightarrow \RR$ where $\xi$ is a discriminative deep network (see Figure~\ref{fig-workflow}, center). This map $h_\xi$ is often referred to as being an ``adversarial'' map. Plugging this ansatz in~\eqref{eq-dual-w} leads to the Wasserstein-GAN problem
\eql{\label{eq-gan}\tag{WGAN}
	\umin{\th} \umax{\xi} \int_{\Zz}  h_\xi \circ g_\xi(z) \d\zeta(z) + \sum_j h_\xi^c(y_j).
}
In the special case where $c(x,y)=\norm{x-y}$, one can prove that the mechanics of $c$-transforms result in the additional constraint that $\tilde h=-h$, subject to $h$ being a $1$-Lipschitz function, see~\cite[Particular case 5.4]{villani2008optimal}. This is used in~\cite{WassersteinGAN} to replace $h_\xi^c$ by $-h_\xi$ in~\eqref{eq-gan} and use a deep network made of ReLu units whose Lipschitz constant is upper-bounded by $1$. 

As a side-note, and as previously commented in the literature, there is at this point no empirical evidence that supports the idea that using discriminative deep networks that way can result in accurate approximations of Wasserstein distances. These alternative formulations provide instead a very useful proxy for a quantity directly related to the Wasserstein distance.

\section{Primal Formulation and VAE}

Following~\cite{Bousquet2017,2017-Genevay-AutoDiff}, in the special case of a generative model $\mu_\th = g_{\th\sharp}\zeta$, formula~\eqref{eq-primal-formula} can be conveniently re-written as
\eql{\label{eq-primal-rewrite}
	E(\th) = \umin{\pi \in \Pp(\Zz \times \Xx)} \enscond{ \int_{\Zz \times \Xx} c(g_\th(z),y) \d \pi(z,y) }{ P_{1\sharp}\pi=\zeta, P_{2\sharp}\pi=\nu }.
}
This is advantageous because now $\pi$ is defined over $\Zz \times \Xx$, which is lower-dimensional than $\Xx \times \Xx$,  and also because, as in Equation~\eqref{eq-dual-w}, $\th$ does not appear in the constraints either.
This provides an alternative formula for the gradient of $E$:
\eql{\label{eq-grad-primal}
	\nabla E(\th) = \int_{\Zz \times \Xx} [\partial_\th g_\th(z)]^\top \nabla_1 c(g_\th(z),y) \d \pi^\star(z,y), 
}
where $\pi^\star$ is an optimal coupling solving~\eqref{eq-primal-rewrite}. Here $\nabla_1 c(x,y) \in \RR^p$ denotes the gradient of $c$ with respect to the first variable.

\cite{Bousquet2017} suggests to look for couplings $\pi$ with a parametric form. A simple way to achieve this is to restrict couplings $\pi$ to those of the form
\eq{
	\pi_{\xi} \eqdef \sum_{j} \de_{(f_\xi(y_j),y_j)} \in \Pp(\Zz \times \Xx),  
}
where $f_\xi : \Xx \rightarrow \Zz$ is a parametric ``encoding'' map (typically a deep network), see Figure~\ref{fig-workflow}, right. This $\pi_\xi$ satisfies by construction the marginal constraint $P_{2\sharp}\pi=\nu$, but in general it cannot satisfy the other constraint $P_{1\sharp}\pi=\zeta$ (because $P_{1\sharp}\pi_\xi$ is discrete while $\zeta$ is not). So following~\cite{Bousquet2017}, it makes sense to consider a relaxed ``unbalanced'' formulation (in the sense of~\cite{2016-chizat-sinkhorn}) of the form
\eql{\label{eq-relaxed-estimator}
	E_\la(\th) = \umin{\pi} \enscond{ \int_{\Zz \times \Xx} c(g_\th(z),y) \d \pi(z,y) + \la D(P_{1\sharp}\pi|\zeta) }{ P_{2\sharp}\pi=\nu }, 
}
where $D(\cdot|\cdot)$ is some distance or divergence between positive measures on $\Zz$ and $\la>0$ a relaxation parameter. 

Plugging the ansatz $\pi=\pi_\xi$ in~\eqref{eq-relaxed-estimator}, one obtains the Wasserstein-VAE formulation
\eql{\label{eq-vae}\tag{WVAE}
	\umin{(\th,\xi)}  \De_\nu( g_\th \circ f_\xi, \Id_{\Xx} )  + \la D( f_{\xi\sharp} \nu |\zeta), 
}
where $\De_\nu( \phi, \Id_{\Xx} )$ is the cost measuring the deviation of a map $\phi : \Xx \rightarrow \Xx$ to identity
\eq{
	\De_\nu( \phi, \Id_{\Xx} ) \eqdef 
	\int_\Xx c(\phi(y),y) \d\nu(y)
	= 
	\frac{1}{n}\sum_{j=1}^n c(\phi(y_j),y_j).
}
Such a cost is usually associated with the Monge formulation of optimal transport~\cite{Monge1781}, whose original motivation was to find an optimal \emph{map} under that cost that would be able to push forward a given measure onto another\cite[\S1.1]{santambrogio2015optimal}.

\section{Conclusions}

\newcommand{\thA}[1]{\th_{\text{\tiny #1}}}

The~\ref{eq-gan} and~\ref{eq-vae} formulations are very different, and are in some sense dual one of each other.  
For GAN, the couple $(g_\th,h_\xi)$ should be thought as a (primal, dual) pair (often referred to as adversarial pair, which is reminiscent of game theory saddle points). For VAE, the couple $(f_\xi,g_\th)$ is rather an (encoding, decoding) pair, and both have the flavour of transportation maps.

In sharp contrast to the primal gradient formula~\eqref{eq-grad-primal} which only requires integrating against an optimal coupling $\pi^\star$, the dual gradient formula~\eqref{eq-grad-dual} involves the integration of the \emph{gradient} of an optimal potential $h^\star$. The latter tends to be more unstable and thus necessitates accurate optimization sub-iterations to obtain an optimal dual potential $h^\star$~\cite{2016-genevay-nips} or an approximation $h_\xi^\star$ within a restricted parametric class~\cite{WassersteinGAN}.  This is somehow inline with the empirical observation that training VAE is more stable than training GAN. 
One should however bear in mind that, although both formulations can be motivated by the same minimum Kantorovitch estimation problem~\eqref{eq-fitting-energy}, they define quite different estimators. In particular, GAN is often credited for producing less blurry outputs when used for image generation. 

Denoting $\thA{MKE}, \thA{WGAN}$ and $\thA{WVAE}$ the solutions of~\eqref{eq-fitting-energy}, \eqref{eq-gan} and \eqref{eq-vae}, one has in the limit $\la \rightarrow +\infty$ (to cancel the bias due to the marginal constraint relaxation), 
\eq{
	E(\thA{WGAN}) \leq E(\thA{MKE}) \leq E(\thA{WVAE}).
}
\cite{Bousquet2017} furthermore mentions that in the ``non-parametric limit'' (i.e. when the number of parameters appearing in $\xi$ tends to $+\infty$, and also letting $\la \rightarrow +\infty$), the gap between the estimators should vanish. Indeed, $h_\xi$ and $f_\xi$ should capture the desired optimal map in the limit and one thus recovers the true solution to~\eqref{eq-fitting-energy}. While it would be interesting from a theoretical perspective to prove and quantify such a claim, it is unclear wether it would be useful for the practitioner. Indeed, the convergence rate might be slow, so that in practice one can be quite far from this non-parametric limit. One could even argue that this limit may give poor estimators for complicated datasets, so that parameterizing the maps and using non-convex optimization solvers lead instead to a beneficial and implicit regularization of these estimators.

\bibliographystyle{plain}
\bibliography{biblio}

\end{document}